%% file: acl2021.tex
\documentclass[11pt,a4paper]{article} 
\usepackage[hyperref]{acl2021}
\usepackage{times}
\usepackage{latexsym}

\usepackage{lipsum}
\newcommand\blfootnote[1]{%
  \begingroup
  \renewcommand\thefootnote{}\footnote{#1}%
  \addtocounter{footnote}{-1}%
  \endgroup
}

\usepackage{graphicx}
\usepackage{multirow}
\usepackage{multicol}
\usepackage{microtype}
\usepackage{subcaption}
\usepackage{enumitem,kantlipsum}
\usepackage[normalem]{ulem}
\usepackage{tabularx}
\usepackage{tcolorbox}
\usepackage{color, colortbl}

\usepackage{pbox}
\usepackage{booktabs}

\newcommand{\rashkincorpus}{\texttt{TSHP-17}}
\newcommand{\proppycorpus}{\texttt{QProp} }

\newcommand{\imgsrc}[2]{\href{#1}{#2}}
\newcommand{\pixelbay}[1]{\href{https://pixabay.com/service/license/}{#1}}
\newcommand{\unplash}[1]{\href{https://unsplash.com/license}{#1}}
\newcommand{\public}[1]{\href{https://creativecommons.org/publicdomain/zero/1.0/}{#1}}
\newcommand{\ccsnd}[1]{\href{https://creativecommons.org/licenses/by-sa/2.0/deed.en}{#1}}
\newcommand{\cctrdunprt}[1]{\href{https://creativecommons.org/licenses/by-sa/3.0/deed.en}{#1}}
\newcommand{\ccfrth}[1]{\href{https://creativecommons.org/licenses/by-sa/4.0/deed.en}{#1}}
\newcommand{\imgur}[1]{\href{https://imgur.com/tos}{#1}}

\aclfinalcopy 
\def\aclpaperid{2503} 

\title{Detecting Propaganda Techniques in Memes}

\author{
Dimitar Dimitrov,\textsuperscript{\rm 1}  
Bishr Bin Ali,\textsuperscript{\rm 2} 
Shaden Shaar,\textsuperscript{\rm 3} 
Firoj Alam,\textsuperscript{\rm 3} 
\\{\bf Fabrizio Silvestri,\textsuperscript{\rm 4}  
Hamed Firooz,\textsuperscript{\rm 5} 
Preslav Nakov, \textsuperscript{\rm 3} 
and Giovanni Da San Martino\textsuperscript{\rm 6} 
}\\
\textsuperscript{\rm 1} Sofia University ``St. Kliment Ohridski'', Bulgaria,
\textsuperscript{\rm 2} King's College London, UK,\\
\textsuperscript{\rm 3} Qatar Computing Research Institute, HBKU, Qatar\\
\textsuperscript{\rm 4} Sapienza University of Rome, Italy,
\textsuperscript{\rm 5} Facebook AI, USA, 
\textsuperscript{\rm 6} University of Padova, Italy\\
\texttt{mitko.bg.ss@gmail.com, bishrkc@gmail.com}\hspace{5mm}\\
\texttt{\{sshaar, fialam, pnakov\}@hbku.edu.qa, mhfirooz@fb.com}\hspace{5mm} \\
\texttt{fsilvestri@diag.uniroma1.it,
dasan@math.unipd.it}
  \\}

\date{}

\begin{document}
\maketitle
\begin{abstract}
Propaganda can be defined as a form of communication that aims to influence the opinions or the actions of people towards a specific goal; this is achieved by means of well-defined rhetorical and psychological devices. Propaganda, in the form we know it today, can be dated back to the beginning of the 17th century. However, it is with the advent of the Internet and the social media that it has started to spread on a much larger scale than before, thus becoming major societal and political issue. Nowadays, a large fraction of propaganda in social media is multimodal, mixing textual with visual content. With this in mind, here we propose a new multi-label multimodal task: detecting the type of propaganda techniques used in memes. We further create and release a new corpus of 950 memes, carefully annotated with 22 propaganda techniques, which can appear in the text, in the image, or in both. Our analysis of the corpus shows that understanding both modalities together is essential for detecting these techniques. This is further confirmed in our experiments with several state-of-the-art multimodal models. 
\end{abstract}

\input{sections/introduction}
\input{sections/related_work}
\input{sections/propaganda_techniques}

\input{sections/dataset}
\input{sections/experiments_models}
\input{sections/results}
\input{sections/conclusion}

\bibliographystyle{acl_natbib}
\bibliography{bib/acl2021,bib/custom,bib/anthology}

\newpage
\clearpage
\section*{Appendix}
\label{sec:appendix}
\appendix
\input{sections/supplemental_material}

\end{document}

%% file: sections/introduction.tex
\section{Introduction}\blfootnote{WARNING: This paper contains meme examples and words that are offensive in nature.}
\label{sec:introduction}

\begin{figure*}[t]
\centering
\includegraphics[width=0.85\textwidth]{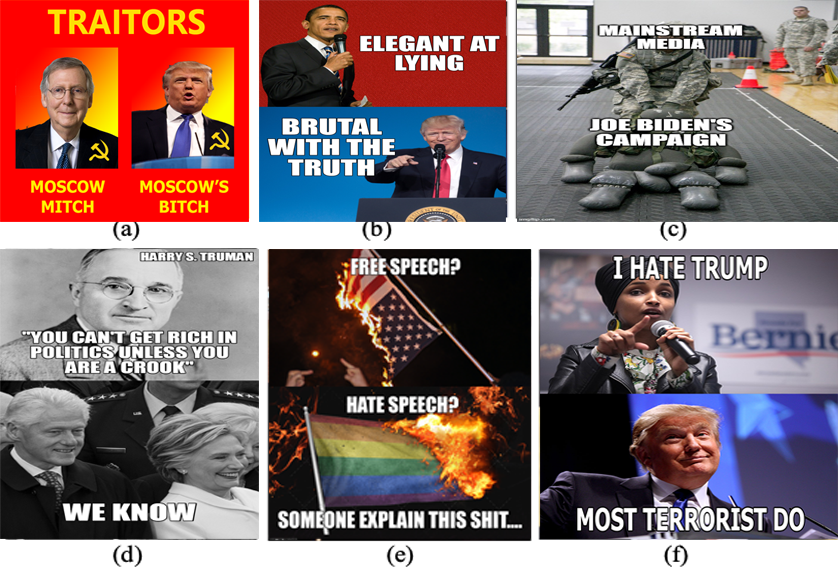}
\caption{Examples of memes from our dataset. \textbf{Image sources:} \imgsrc{https://commons.wikimedia.org/wiki/File:Sen_Mitch_McConnell_official_(cropped).jpg}{(a) 1}, 
    \imgsrc{https://search.creativecommons.org/photos/ce37cc08-cf83-4a5b-9903-80d9e7d172e7}{(a) 2}; \imgsrc{https://upload.wikimedia.org/wikipedia/commons/1/1b/Barack_Obama_February_2_2008.jpg}{(b) 1},
    \imgsrc{https://upload.wikimedia.org/wikipedia/commons/9/9c/Donald_Trump_\%2832984318012\%29.jpg}{(b) 2}; \imgsrc{https://upload.wikimedia.org/wikipedia/commons/2/24/One_team_one_fight\%2C_1st_Cav_participates_in_new_study_140717-A-TK117-157.jpg}{(c)};
    \imgsrc{https://images.unsplash.com/photo-1580128636867-7224f71904fd?ixid=MXwxMjA3fDB8MHxwaG90by1wYWdlfHx8fGVufDB8fHw\%3D&ixlib=rb-1.2.1&auto=format&fit=crop&w=696&q=80}{(d) 1},
    \imgsrc{https://commons.wikimedia.org/wiki/File:Bill_and_Hillary_Clinton_at_58th_Inauguration_01-20-17_(cropped).jpg}{(d) 2};
    \imgsrc{https://upload.wikimedia.org/wikipedia/commons/1/1f/Shaw_Day_2_Photo_18.jpg}{(e) 1},
    \imgsrc{https://pixabay.com/illustrations/flag-lgbt-gay-lgbtq-lesbian-1184117/}{(e) 2};
    \imgsrc{https://commons.wikimedia.org/wiki/File:Ilhan_Omar_(49518038586).jpg}{(f) 1},
    \imgsrc{https://commons.wikimedia.org/wiki/File:Donald_Trump_by_Gage_Skidmore_2.jpg}{(f) 2}; \textbf{Licenses:} \public{(a) 1, (c), (d) 2}; \ccsnd{(a) 2, (b) 2, (f) 1}; \cctrdunprt{(b) 1, (f) 2}; \unplash{(d) 1}; \ccfrth{(e) 1}; \pixelbay{(e) 2}
}
\label{fig:meme_example}
\end{figure*}

Social media have become one of the main communication channels for information dissemination and consumption, and nowadays many people rely on them as their primary source of news~\cite{perrin2015social}. Despite the many benefits that social media offer, sporadically they are also used as a tool, by bots or human operators, to manipulate and to mislead unsuspecting users. Propaganda is one such communication tool to influence the opinions and the actions of other people in order to achieve a predetermined goal~\cite{InstituteforPropagandaAnalysis1938}. 

Propaganda is not new. It can be traced back to the beginning of the 17th century, as reported in \cite{Margolin1979TheVR,pamphlet_casey,da2020survey}, where the manipulation was present at public events such as theaters, festivals, and during games. In the current information ecosystem, it has evolved to \textit{computational propaganda} \cite{woolley2018computational,da2020survey}, where information is distributed through technological means to social media platforms, which in turn make it possible to reach well-targeted communities at high velocity. We believe that being aware and able to detect propaganda campaigns would contribute to a healthier online environment. 

Propaganda appears in various forms and has been studied by different research communities. 
There has been work on exploring network structure, looking for malicious accounts and coordinated inauthentic behavior~\cite{10.1145/3041021.3055135,yang2019arming,chetan2019corerank,pacheco2020unveiling}. 

In the natural language processing community, propaganda has been studied at the document level~\cite{BARRONCEDENO20191849,rashkin-EtAl:2017:EMNLP2017}, and at the sentence and the fragment levels~\cite{EMNLP19DaSanMartino}. 
There have also been notable datasets developed, including (\emph{i})~TSHP-17 \cite{rashkin-EtAl:2017:EMNLP2017}, which consists of document-level annotation labeled with four classes (trusted, satire, hoax, and propaganda); (\emph{ii})~QProp \cite{BARRONCEDENO20191849}, which uses binary labels (propaganda vs. non-propaganda), and (\emph{iii})~PTC \cite{EMNLP19DaSanMartino}, which uses fragment-level annotation and an inventory of 18 propaganda techniques. While that work has focused on text, here we aim to detect propaganda techniques from a multimodal perspective. This is a new research direction, even though large part of propagandistic social media content nowadays is multimodal, e.g.,~in the form of memes. 

Memes are popular in social media as they can be quickly understood with minimal effort \cite{diresta2018computational}. They can easily become viral, and thus it is important to detect malicious ones quickly, and also to understand the nature of propaganda, which can help human moderators, but also journalists, by offering them support for a higher level analysis.

Figure~\ref{fig:meme_example} shows some examples of memes\footnote{In order to avoid potential copyright issues, all memes we show in this paper are our own recreation of existing memes, using images with clear licenses.} and propaganda techniques. Example \textit{\textbf{(a)}} applies \emph{transfer}, using symbols (hammer and sickle) and colors (red), that are commonly associated with communism, in relation to the two Republicans shown in the image; it also uses \emph{Name Calling} (\textit{traitors}, \emph{Moscow Mitch}, \emph{Moscow's bitch}). The meme in \textit{\textbf{(b)}} uses both \textit{Smears} and \textit{Glittering Generalities}. The one in \textit{\textbf{(c)}} expresses \textit{Smears} and suggest that Joe Biden's campaign is only alive because of mainstream media. The examples in the second row show some less common techniques. Example \textbf{\textit{(d)}} uses \textit{Appeal to authority} to give credibility to a statement that rich politicians are crooks, and there is also a \textit{Thought-terminating clich\'{e}} used to discourage critical thought on the statement in the form of the phrase \textit{``WE KNOW''}, thus implying that the Clintons are crooks, which is also \emph{Smears}.

Then, example \textbf{\textit{(e)}} uses both \textit{Appeal to (Strong) Emotions} and \textit{Flag-waving} as it tries to play on patriotic feelings. Finally, example \textbf{\textit{(f)}} has \textit{Reduction ad hitlerum} as Ilhan Omars' actions are related to such of a terrorist (which is also \emph{Smears}; moreover, the word \textit{HATE} expresses \textit{Loaded language}). 

The above examples illustrate that propaganda techniques express shortcuts in the argumentation process, e.g.,~by leveraging on the emotions of the audience or by using logical fallacies to influence it. Their presence does not necessarily imply that the meme is propagandistic. Thus, we do not annotate whether a meme is propagandistic (just the propaganda techniques it contains), as this would require, among other things, to determine its intent.

Our contributions can be summarized as follows:
\begin{itemize}
    \item We formulate a new multimodal task: propaganda detection in memes, and we discuss how it relates and differs from previous work.
    \item We develop a multi-modal annotation schema, and we create and release a new dataset for the task, consisting of 950 memes, which we manually annotate with 22 propaganda techniques.\footnote{The corpus and the code used in our experiments are available at \url{ https://github.com/di-dimitrov/propaganda-techniques-in-memes}.}
    \item We perform manual analysis, and we show that both modalities (text and images) are important for the task.
    \item We experiment with several state-of-the-art textual, visual, and multimodal models, which further confirm the importance of both modalities, as well as the need for further research.
\end{itemize}

%% file: sections/related_work.tex
\section{Related Work}
\label{sec:related_work}

\paragraph{Computational Propaganda}

Computational propaganda is defined as the use of automatic approaches to intentionally disseminate misleading information over social media platforms \cite{woolley2018computational}. The information that is distributed over these channels can be textual, visual, or multi-modal. Of particular importance are memes, which can be quite effective at spreading multimodal propaganda on social media platforms \cite{diresta2018computational}. The current information ecosystem and virality tools, such as bots, enable memes to spread easily, jumping from one target group to another. As of present, attempts to limit the spread of such memes have focused on analyzing social networks and looking for fake accounts and bots to reduce the spread of such content \cite{10.1145/3041021.3055135,yang2019arming,chetan2019corerank,pacheco2020unveiling}.

\paragraph{Textual Content}
Most research on propaganda detection has focused on analyzing textual content \cite{BARRONCEDENO20191849,rashkin-EtAl:2017:EMNLP2017,EMNLP19DaSanMartino,da2020survey}. 
\citet{rashkin-EtAl:2017:EMNLP2017} developed the \rashkincorpus corpus, which uses document-level annotation and is labeled with four classes: \emph{trusted}, \emph{satire}, \emph{hoax}, and \emph{propaganda}. \rashkincorpus~ was developed using distant supervision, i.e.,~all articles from a given news outlet share the label of that outlet. The articles were collected from the English Gigaword corpus and from seven other unreliable news sources. Among them two were propagandistic. They trained a model using word $n$-gram representation with logistic regression and reported that the model performed well only on articles from sources that the system was trained on.

\citet{BARRONCEDENO20191849} developed a new corpus, \proppycorpus, with two labels: propaganda vs. non-propaganda. They also experimented on \rashkincorpus~and \proppycorpus~corpora, where for the \rashkincorpus~ corpus, they binarized the labels: propaganda \textit{vs.} any of the other three categories. 

They performed massive experiments, investigated writing style and readability level, and trained models using logistic regression and SVMs. Their findings confirmed that using distant supervision, in conjunction with rich representations, might encourage the model to predict the source of the article, rather than to discriminate propaganda from non-propaganda. 
Similarly, \citet{Habernal.et.al.2017.EMNLP,Habernal2018b} developed a corpus with 1.3k arguments annotated with five fallacies, including \textit{ad hominem}, \textit{red herring}, and \textit{irrelevant authority}, which directly relate to propaganda techniques.

A more fine-grained propaganda analysis was done by \citet{EMNLP19DaSanMartino}. They developed a corpus of news articles annotated with 18 propaganda techniques. The annotation was at the fragment level, and enabled two tasks: (\emph{i})~binary classification ---given a sentence in an article, predict whether any of the 18 techniques has been used in it; (\emph{ii})~multi-label multi-class classification and span detection task ---given a raw text, identify both the specific text fragments where a propaganda technique is being used as well as the type of the technique. On top of this work, they proposed a multi-granular deep neural network that captures signals from the sentence-level task and helps to improve the fragment-level classifier. Subsequently, a system was developed and made publicly available \cite{da2020prta}.

\paragraph{Multimodal Content}
Previous work has explored the use of multimodal content for detecting misleading information \cite{Volkova_Ayton_Arendt_Huang_Hutchinson_2019}, deception \cite{Glenski2019MultilingualMD}, emotions and propaganda \cite{abd_kadir_etal}, hateful memes \cite{kiela2020hateful,lippe2020multimodal,das2020detecting}, antisemitism \cite{chandra2021subverting} and propaganda in images \cite{doi:10.1080/15551393.2014.955501}. 
\citet{Volkova_Ayton_Arendt_Huang_Hutchinson_2019} proposed models for detecting misleading information using images and text. They developed a corpus of 500,000 Twitter posts consisting of images labeled with six classes: disinformation, propaganda, hoaxes, conspiracies, clickbait, and satire. 
Then, they modeled textual, visual, and lexical characteristics of the text.
\citet{Glenski2019MultilingualMD} explored multilingual multimodal content for deception detection. They had two multi-class classification tasks: (\emph{i})~classifying social media posts into four categories (propaganda, conspiracy, hoax, or clickbait), and (\emph{ii})~classifying social media posts into five categories (disinformation, propaganda, conspiracy, hoax, or clickbait). 

Multimodal hateful memes have been the target of the popular ``Hateful Memes Challenge'', which the participants addressed using fine-tuned state-of-art multi-modal transformer models such as ViLBERT \cite{lu2019vilbert}, Multimodal Bitransformers \cite{kiela2019supervised}, and VisualBERT \cite{li2019visualbert} to classify hateful vs. not-hateful memes \cite{kiela2020hateful}. \citet{lippe2020multimodal} explored different early-fusion multimodal approaches and proposed various methods that can improve the performance of the detection systems. 

Our work differs from the above research in terms of annotation, as we have a rich inventory of 22 fine-grained propaganda techniques, which we annotate separately in the text and then jointly in the text+image, thus enabling interesting analysis as well as systems for multi-modal propaganda detection with explainability capabilities.

%% file: sections/propaganda_techniques.tex
\section{Propaganda Techniques}
\label{sec:annotation}

Propaganda comes in many forms and over time a number of techniques have emerged in the literature \cite{Torok2015,Miller,EMNLP19DaSanMartino,web_smear,article_kadir,book_prop,web_hobbs}. Different authors have proposed inventories of propaganda techniques of various sizes: seven techniques \cite{Miller}, 24 techniques \citet{Weston2000}, 18 techniques \cite{EMNLP19DaSanMartino}, just smear as a technique \cite{web_smear}, and seven techniques \cite{article_kadir}. We adapted the techniques discussed in \cite{EMNLP19DaSanMartino}, \cite{web_smear} and \cite{article_kadir}, thus ending up with 22 propaganda techniques. Among our 22 techniques, the first 20 are used for both text and images, while the last two \textit{Appeal to (Strong) Emotions} and \textit{Transfer} are reserved for labeling images only. Below, we provide the definitions of these techniques, which are included in the guidelines the annotators followed (see appendix~\ref{ssec:appendix_guideline_phase2-5}) for more detal.

\begin{enumerate}[leftmargin=*]
   \itemsep0em
    \item \textbf{Loaded language:} Using specific words and phrases with strong emotional implications (either positive or negative) to influence an audience.
    \item \textbf{Name calling or labeling:} Labeling the object of the propaganda campaign as something that the target audience fears, hates, finds undesirable or loves, praises.
    \item \textbf{Doubt:} Questioning the credibility of someone or something.
    \item \textbf{Exaggeration / Minimisation:} Either representing something in an excessive manner: making things larger, better, worse (e.g., \emph{the best of the best}, \emph{quality guaranteed}) or making something seem less important or smaller than it really is (e.g.,~saying that an insult was actually just a joke).
    \item \textbf{Appeal to fear / prejudices:} Seeking to build support for an idea by instilling anxiety and/or panic in the population towards an alternative. In some cases, the support is built based on preconceived judgements.
    \item \textbf{Slogans:} A brief and striking phrase that may include labeling and stereotyping. Slogans tend to act as emotional appeals.
    \item \textbf{Whataboutism:} A technique that attempts to discredit an opponent's position by charging them with hypocrisy without directly disproving their argument.
    \item \textbf{Flag-waving:} Playing on strong national feeling (or to any group; e.g.,~race, gender, political preference) to justify or promote an action or an idea.
    \item \textbf{Misrepresentation of someone's position (Straw man):} Substituting an opponent's proposition with a similar one, which is then refuted in place of the original proposition.
    \item \textbf{Causal oversimplification:} Assuming a single cause or reason when there are actually multiple causes for an issue. This includes transferring blame to one person or group of people without investigating the complexities of the issue.
    \item \textbf{Appeal to authority:} Stating that a claim is true simply because a valid authority or expert on the issue said it was true, without any other supporting evidence offered. We also include here the special case where the reference is not an authority or an expert, which is referred to as \emph{Testimonial} in the literature.
    \item \textbf{Thought-terminating clich\'{e}:} Words or phrases that discourage critical thought and meaningful discussion about a given topic. They are typically short, generic sentences that offer seemingly simple answers to complex questions or that distract the attention away from other lines of thought.
    \item \textbf{Black-and-white fallacy or dictatorship:} Presenting two alternative options as the only possibilities, when in fact more possibilities exist. As an the extreme case, tell the audience exactly what actions to take, eliminating any other possible choices (Dictatorship).
    \item \textbf{Reductio ad hitlerum:} Persuading an audience to disapprove an action or an idea by suggesting that the idea is popular with groups hated in contempt by the target audience. It can refer to any person or concept with a negative connotation.
    \item \textbf{Repetition:} Repeating the same message over and over again, so that the audience will eventually accept it.
    \item \textbf{Obfuscation, Intentional vagueness, Confusion:} Using words that are deliberately not clear, so that the audience may have their own interpretations. For example, when an unclear phrase with multiple possible meanings is used within an argument and, therefore, it does not support the conclusion.
    \item \textbf{Presenting irrelevant data (Red Herring):} Introducing irrelevant material to the issue being discussed, so that everyone's attention is diverted away from the points made.
    \item \textbf{Bandwagon} Attempting to persuade the target audience to join in and take the course of action because ``everyone else is taking the same action.''
    \item \textbf{Smears:} A smear is an effort to damage or call into question someone's reputation, by propounding negative propaganda. It can be applied to individuals or groups.
    \item \textbf{Glittering generalities (Virtue):} These are words or symbols in the value system of the target audience that produce a positive image when attached to a person or an issue.
    \item \textbf{Appeal to (strong) emotions:} Using images with strong positive/negative emotional implications to influence an audience.
    \item \textbf{Transfer:} Also known as \emph{association}, this is a technique that evokes an emotional response by projecting positive or negative qualities (praise or blame) of a person, entity, object, or value onto another one in order to make the latter more acceptable or to discredit it.
\end{enumerate}

%% file: sections/dataset.tex
\begin{figure*}[tbh]
\centering
\includegraphics[width=0.95\textwidth]{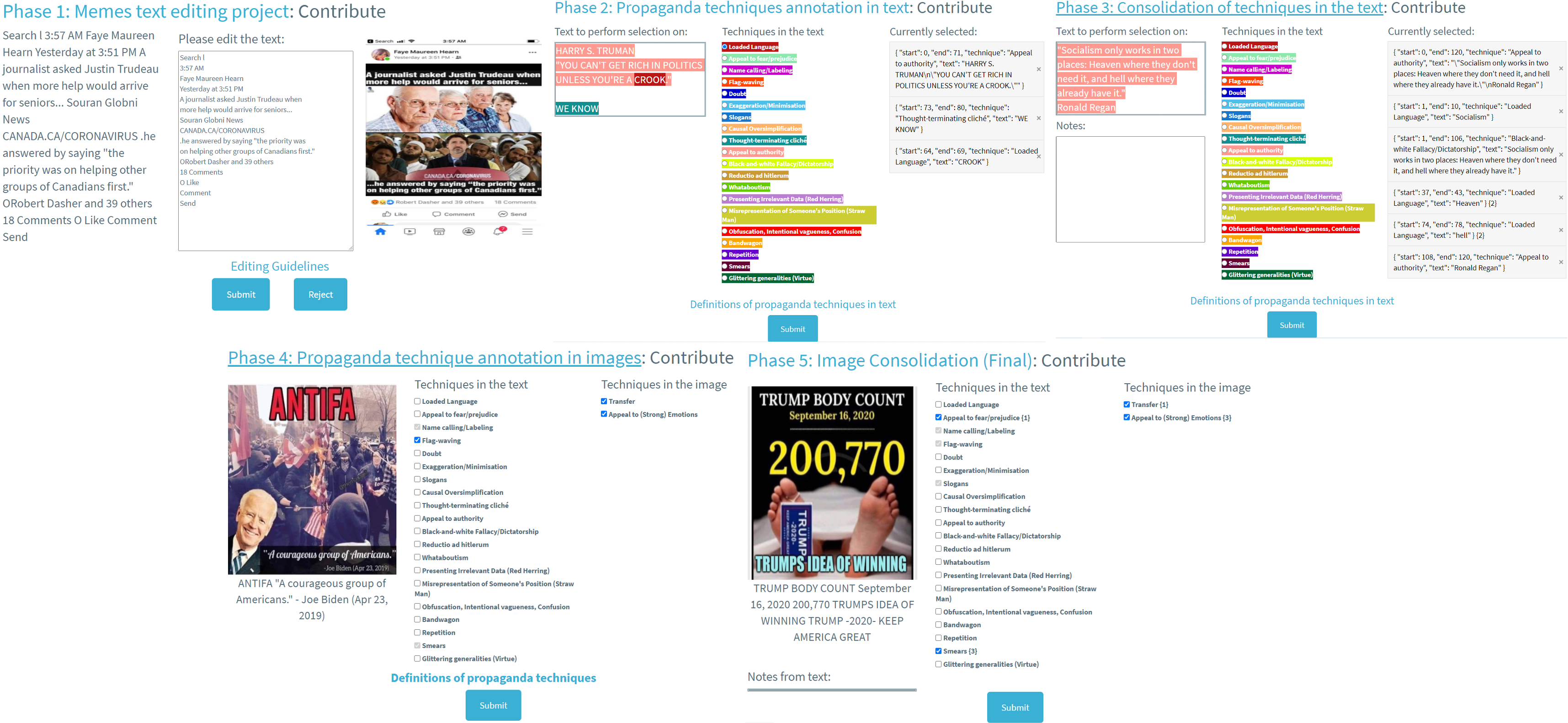}
\caption{Examples of the annotation interface of each phase of the annotation process.}
\label{fig:pyb_example}
\end{figure*}

\section{Dataset}
\label{sec:dataset}

We collected memes from our own private Facebook accounts, and we followed various Facebook public groups on different topics such as vaccines, politics (from different parts of the political spectrum), COVID-19, gender equality, and more. We wanted to make sure that we have a constant stream of memes in the newsfeed. We extracted memes at different time frames, i.e.,~once every few days for a period of three months. We also collected some old memes for each group in order to make sure we covered a larger variety of topics.

\subsection{Annotation Process}
We annotated the memes using the 22 propaganda techniques described in Section~\ref{sec:annotation} in a multilabel setup. The motivation for multilabel annotation is that the content in the memes often expresses multiple techniques, even though such a setting adds complexity both in terms of annotation and of classification. We also chose to consider annotating spans because the propaganda techniques can appear in the different chunk(s), which is also in line with recent research \cite{EMNLP19DaSanMartino}. We could not consider annotating the visual modality independently because all memes contain the text as part of the image.

The annotation team included six members, both female and male, all fluent in English, with qualifications ranging from undergrad, to MSc and PhD degrees, including experienced NLP researchers; this helped to ensure the quality of the annotation. No incentives were provided to the annotators. The annotation process required understanding the textual and the visual content, which poses a great challenge for the annotator. Thus, we divided it into five phases, as discussed below and as shown in Figure \ref{fig:pyb_example}. Among these phases there were three stages, (\emph{i})~pilot annotations to train the annotators to recognize the propaganda techniques, (\emph{ii})~independent annotations by three annotators for each meme (phase 2 and 4), (\emph{iii})~consolidation (phase 3 and 5), where the annotators met with the other three team members, who acted as consolidators, and all six discussed every single example in detail (even those for which there was no disagreement).

We chose PyBossa\footnote{\url{https://pybossa.com}} as our annotation platform as it provides the functionality to create a custom annotation interface that can fit our needs in each phase of the annotation.

\subsubsection{Phase 1: Filtering and Text Editing}

Phase 1 is about filtering some of the memes according to our guidelines, e.g.,~low-quality memes, and such containing no propaganda technique.
We automatically extracted the textual content using OCR, and then post-edited it to correct for potential OCR errors. We filtered and edited the text manually, whereas for extracting the text, we used the Google Vision API.\footnote{\url{http://cloud.google.com/vision}} We presented the original meme and the extracted text to an annotator, who had to filter and to edit the text in phase 1 as shown in Figure~\ref{fig:pyb_example}. For filtering and editing, we defined a set of rules, e.g.,~we removed hard to understand, or low-quality images, cartoons, memes with no picture, no text, or for which the textual content was strongly dominant and the visual content was minimal and uninformative, e.g.,~a single-color background. More details about filtering and editing are given in Appendix \ref{apendix:rejection} and \ref{apendix:editing}.

\subsubsection{Phase 2: Text Annotation}
In phase 2, we presented the edited textual content of the meme to the annotators as shown in Figure~\ref{fig:pyb_example}. We asked the annotators to identify the propaganda techniques in the text and to select the corresponding text spans for each of them.

\subsubsection{Phase 3: Text Consolidation}
Phase 3 is the consolidation step of the annotations from phase 2 as shown in Figure~\ref{fig:pyb_example}. This phase was essential for ensuring the quality, and it further served as an additional training opportunity for the entire team, which we found very useful.

\subsubsection{Phase 4: Multimodal Annotation}
Step 4 is multimodal meme annotation, i.e.,~considering both the textual and the visual content in the meme. In this phase, we show the meme, the post-edited text, and the consolidated propaganda labels from phase 3 (text only) to the annotators, as shown in phase 4 from Figure~\ref{fig:pyb_example}.
We intentionally provided the consolidated text labels to the annotators in this phase because we wanted them to focus on the techniques that require the presence of the image rather than to reannotate those from the text.\footnote{Ideally, we would have wanted to have also a phase to annotate propaganda techniques when showing the image only; however, this is hard to do in practice as the text is embedded as part of the pixels in the image.}

\subsubsection{Phase 5: Multimodal Consolidation}
This is the consolidation phase for Phase 4; the setup is like for the consolidation at Phase 3, as shown in Figure~\ref{fig:pyb_example}.

Note that, in the majority of the cases, the main reason why two annotations of the same meme might differ was due to one of the annotators not spotting some of the techniques, rather than because there was a disagreement on what technique should be chosen for a given textual span or what the exact boundaries of the span for a given technique instance should be. In the rare cases in which there was an actual disagreement and no clear conclusion could be reached during the discussion phase, we resorted to discarding the meme (there were five such cases in total).

\subsection{Quality of the Annotations} We assessed the quality of the annotations for the individual annotators from phases 2 and 4 (thus, combining the annotations for text and images) to the final consolidated labels at phase 5, following the setting in~\cite{EMNLP19DaSanMartino}.
Since our annotation is multilabel, we computed Krippendorff's $\alpha$, which supports multi-label agreement computation \cite{artstein2008inter,passonneau2006measuring}. 
The results are shown in Table~\ref{tab:annotation_agr} and indicate moderate to perfect agreement \cite{landis1977measurement}.

\begin{table}[h]
\centering
\scalebox{0.90}{
\begin{tabular}{lc}
\toprule
\bf Agreement Pair & \bf Krippendorff's $\alpha$ \\
\midrule
Annotator 1 vs. Consolidated & 0.83 \\
Annotator 2 vs. Consolidated & 0.91 \\
Annotator 3 vs. Consolidated & 0.56 \\
\midrule
\textbf{Average} & \textbf{0.77} \\
\bottomrule
\end{tabular}
}
\caption{Inter-annotator agreement.}
\label{tab:annotation_agr}
\end{table}

\begin{table}[tbh]
\centering
\setlength{\tabcolsep}{2.5pt}
\scalebox{0.8}{
 \begin{tabular}{lrrr}
 \toprule
 \textbf{Propaganda Techniques}  &\multicolumn{2}{c}{\bf Text-Only}  & \textbf{Meme}  \\
  &  \textbf{Len.} & \textbf{\#} & \textbf{\#}  \\
  \midrule
Loaded Language   & 2.41 & 761  & 492  \\
Name calling/Labeling    & 2.62  & 408  &347  \\
Smears   & 17.11  & 266 & 602  \\
Doubt     & 13.71 &  86  &111 \\
Exaggeration/Minimisation    & 6.69 &  85  &  100\\
Slogans    & 4.70 &  72  & 70 \\
Appeal to fear/prejudice   & 10.12 &  60  & 91 \\
Whataboutism    & 22.83 &  54  & 67\\
Glittering generalities (Virtue)  & 14.07 &  45   &112  \\
Flag-waving    & 5.18 &  44 &  55 \\
Repetition & 1.95 &  42  & 14   \\ 
Causal Oversimplification  & 14.48 &  33   &  36 \\
Thought-terminating clich\'{e}   & 4.07 &  28  & 27  \\
Black-and-white Fallacy/Dictatorship  & 11.92 &  25   & 26\\
Straw Man & 15.96 &  24  & 40 \\
Appeal to authority  & 20.05 &  22    & 35 \\
Reductio ad hitlerum    & 12.69 &  13  & 23  \\
Obfuscation, Int. vagueness, Confusion &  9.8 &  5 & 7\\
Presenting Irrelevant Data  &     15.4 &   5  & 7 \\
Bandwagon  &   8.4 &   5   & 5    \\
\hline
Transfer & --- &  --- & 95  \\
Appeal to (Strong) Emotions & --- & --- & 90  \\
\midrule
\textbf{Total} & & \textbf{2,119}  & \textbf{2,488} \\
 \bottomrule
\end{tabular}
}
\caption{Statistics about the propaganda techniques. For each technique, we show the average length of its span (in number of words) as well as the number of instances of the technique as annotated in the text only vs. annotated in the entire meme.}
\label{tab:instances_tasks}
\end{table}

\begin{figure}[tbh]
\centering
\includegraphics[width=0.47\textwidth]{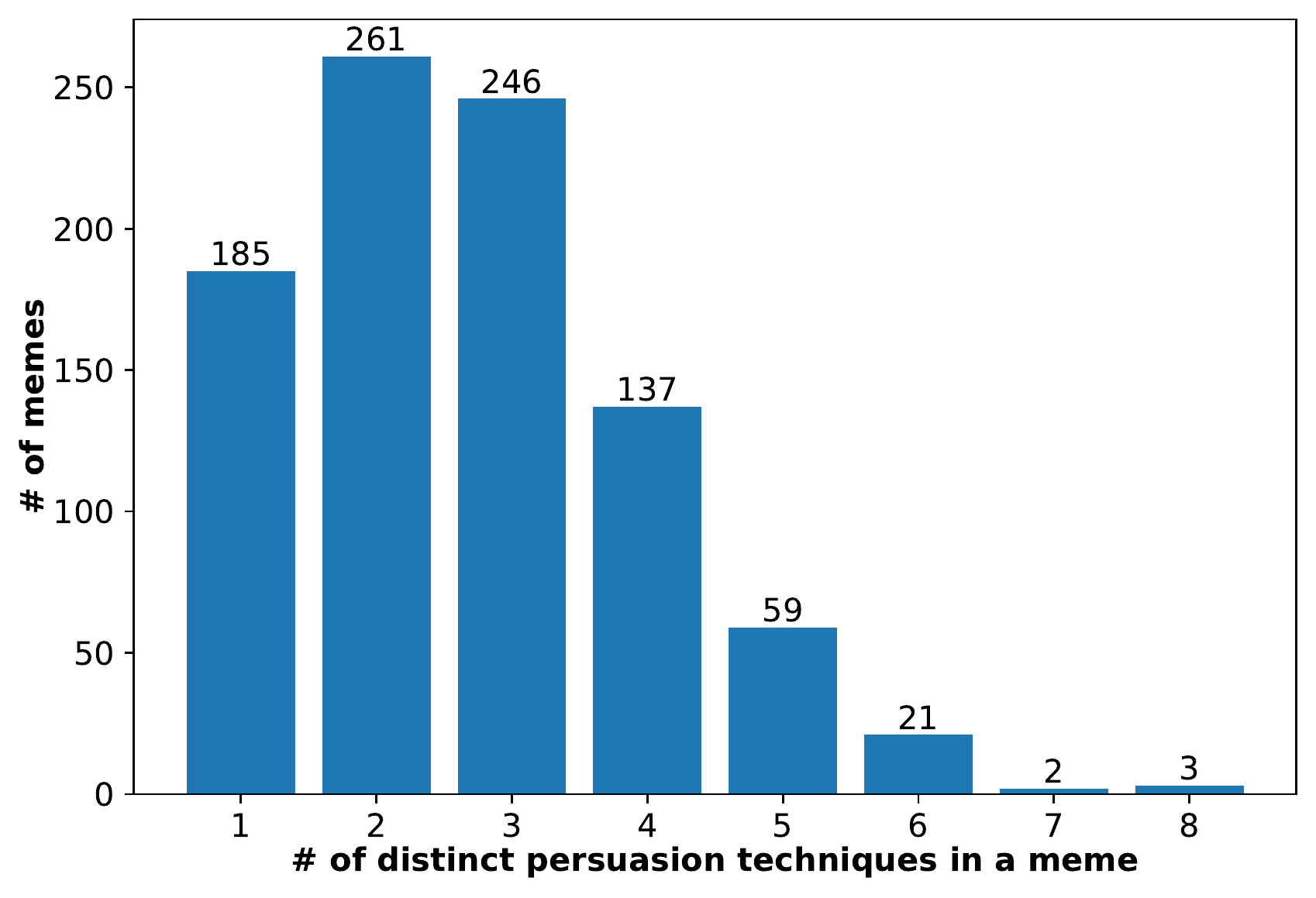}
\caption{Propaganda techniques per meme.}
\label{fig:label_disttext}
\end{figure}

\subsection{Statistics}

After the filtering in phase 1 and the final consolidation, our dataset consists of 950 memes. The maximum number of sentences per meme is 13, but most memes comprise only very few sentences, with an average of 1.68. The number of words ranges between 1 and 73 words, with an average of 17.79$\pm$11.60.
In our analysis, we observed that some propaganda techniques were more textual, e.g.,~\textit{Loaded Language} and \emph{Name Calling}, while others, such as \textit{Transfer}, tended to be more image-related. 

Table~\ref{tab:instances_tasks} shows the number of instances of each technique when using unimodal (text only, i.e.,~after phase~3) vs. multimodal (text + image, i.e.,~after phase~5) annotations. Note also that a total of 36 memes had no propaganda technique annotated. We can see that the most common techniques are \emph{Smears}, \emph{Loaded Language}, and \emph{Name calling/Labeling}, covering 63\%, 51\%, and 36\% of the examples, respectively.
These three techniques also form the most common pairs and triples in the dataset as shown in Table~\ref{tab:multiples}.
We further show the distribution of the number of propaganda techniques per meme in Figure~\ref{fig:label_disttext}. We can see that most memes contain more than one technique, with a maximum of 8 and an average of 2.61. 

Table~\ref{tab:instances_tasks} shows that the techniques can be found both in the textual and in the visual content of the meme, thus suggesting the use of multimodal learning approaches to effectively exploit all information available. Note also that different techniques have different span lengths. For example, \emph{Loaded Language} is about two words long, e.g.,~\textit{violence}, \textit{mass shooter}, and \textit{coward}. However, techniques such as \textit{Whataboutism} need much longer spans with an average length of 22 words.

\begin{table}[tbh]
\scalebox{0.80}{
 \begin{tabular}{l l r} 
 \toprule
 &\textbf{Most common} & \textbf{Freq.} \\
 \midrule
  \multirow{5}{*}{\bf Triples}& loaded lang., name call./labeling, smears & 178 \\
  & name call./labeling, smears, transfer & 40 \\
  & loaded language, smears, transfer & 33 \\
  & appeal to emotions, loaded lang., smears & 30 \\
  & exagg./minim., loaded lang., smears & 28 \\
  \midrule
  \multirow{5}{*}{\bf Pairs}& loaded language, smears & 309 \\
  & name calling/labeling, smears & 256 \\
  & loaded language, name calling/labeling & 243 \\
  & smears, transfer & 75 \\
  & exaggeration/minimisation, smears & 63 \\
 \bottomrule
\end{tabular}
}
\caption{The five \textbf{most common} triples and pairs in our corpus and their \textbf{frequency}.}
\label{tab:multiples}
\end{table}

%% file: sections/experiments_models.tex
\section{Experiments}
\label{sec:exp_setup}

Among the learning tasks that can be defined on our corpus, here we focus on the following one: given a meme, find all the propaganda techniques used in it, both in the text and in the image, i.e.,~predict the techniques as per phase 5.

\subsection{Models}\label{sec:models}

We used two na\"{i}ve baselines.
First, a \textit{Random} baseline, where we assign a technique uniformly at random. Second, a \textit{Majority class} baseline, which always predicts the most frequent class: \textit{Smears}.

\paragraph{Unimodal: text only.} For the text-based unimodal experiments, we used BERT \cite{text_bert}, which is a state-of-the-art pre-trained Transformer, and fastText \cite{joulin2017bag}, which can tolerate potentially noisy text from social media as it is trained on word and character $n$-grams.

\paragraph{Unimodal: image.} For the image-based unimodal experiments, we used ResNet152 \cite{he2016deep}, which was successfully applied in a related setup \cite{kiela2019supervised}.

\paragraph{Multimodal: unimodally pretrained} For the multimodal experiments, we trained separate models on the text and on the image, BERT and ResNet-152, respectively, and then we combined them using (a)~early fusion Multimodal Bitransformers (MMBT) \cite{kiela2019supervised}, (b)~middle fusion (feature concatenation), and (c)~late fusion (combining the predictions of the models). For middle fusion, we took the output of the second-to-last layer of ResNet-152 for the visual part and the output of the \texttt{[CLS]} token from BERT, and we fed them into a multilayer network.

\paragraph{Multimodal: joint models.} We further experimented with models trained using a multimodal objective. In particular, we used ViLBERT \cite{lu2019vilbert}, which is pretrained on Conceptual Captions \cite{sharma2018conceptual}, and Visual BERT \cite{lin2014microsoft}, which is pretrained on the MS-COCO dataset \cite{lin2014microsoft}.

\subsection{Experimental Settings}

We split the data into training, development, and testing with 687 (72\%), 63 (7\%), and 200 (21\%) examples, respectively. Since we are dealing with a multi-class multi-label task, where the labels are imbalanced, we chose micro-average F$_1$ as our main evaluation measure, but we also report macro-average F$_1$.

We used the Multimodal Framework (MMF) \cite{singh2020mmf}. We trained all models on Tesla P100-PCIE-16GB GPU with the following manually tuned hyper-parameters (on dev): batch size of 32, early stopping on the validation set optimizing for F1-micro, sequence length of 128, AdamW as an optimizer with learning rate of 5e-5, epsilon of 1e-8, and weight decay of 0.01. All reported results are averaged over three runs with random seeds. The average execution time for BERT was 30 minutes, and for the other models it was 55 minutes.

%% file: sections/results.tex
\begin{table*}[tbh]
\centering
\small
 \begin{tabular}{l c l r r} 
 \toprule
 \bf Type & \bf \# &\bf Model & \bf F1-Micro &  \bf F1-Macro  \\ [0.5ex] 
  \midrule
 \multirow{2}{*}{Baseline}&1& Random & 7.06 & 5.15\\
  &2& Majority class & 29.04 & 3.12 \\
 \midrule
 \multirow{3}{*}{Unimodal}&3& ResNet-152 & 29.92 & 7.99 \\
  &4& FastText  & 33.56 & 5.25 \\
  &5& BERT & 37.71 & 15.62 \\
  \midrule
  \multirow{6}{*}{Multimodal} & 6& BERT + ResNet-152 (late fusion) & 30.73 & 1.14 \\
  &7& FastText + ResNet-152 (mid-fusion) & 36.12 & 7.52 \\
  &8& BERT + ResNet-152 (mid-fusion) & 38.12 & 10.56 \\
  &9& MMBT & 44.23 & 8.31\\
  &10& ViLBERT CC & 46.76 & 8.99 \\
  &11& VisualBERT COCO & 48.34  & 11.87 \\
 \bottomrule
\end{tabular}
\caption{Evaluation results.}
\label{tab:results_compariosn}
\end{table*}

\section{Experimental Results}
\label{sec:results}

Table \ref{tab:results_compariosn} shows the results for the models in Section~\ref{sec:models}. Rows 1 and 2 show a random and a majority class baseline, respectively. Rows 3-5 show the results for the unimodal models. While they all outperform the baselines, we can see that the model based on visual modality only, i.e.,~ResNet-152 (row 3), performs worse than models based on text only (rows 4-5).
This might indicate that identifying the techniques in the visual content is a harder task than in texts. 
Moreover, BERT significantly outperforms fastText, which is to be expected as it can capture contextual representation better.

Rows 6-8 present results for multimodal fusion models. The best one is BERT + ResNet-152 (+2 points over fastText + ResNet-152).  
We observe that early fusion models (rows 7-8) outperform late fusion ones (row 6). This makes sense as late fusion is a simple mean of the results of each modality, while early fusion has a more complex architecture and trains a separate multi-layer perceptron for the visual and for the textual features. 

We can also see that both mid-fusion models (rows 7-8) improve over the corresponding text-only ones (rows 3-5).
Finally, looking at the results in rows 9-11, we can see that each multimodal model consistently outperforms each of the unimodal models (rows 1-8).
The best results are achieved with ViLBERT CC (row 10) and VisualBERT COCO (row 11), which use complex representations that combine the textual and the visual modalities. Overall, we can conclude that multimodal approaches are necessary to detect the use of propaganda techniques in memes, and that pretrained transformer models seem to be the most promising approach. 

%% file: sections/conclusion.tex
\section{Conclusion and Future Work}
\label{sec:conclutions}

We have proposed a new multi-class multi-label multimodal task: detecting the type of propaganda techniques used in memes. We further created and released a corpus of 950 memes annotated with 22 propaganda techniques, which can appear in the text, in the image, or in both. Our analysis of the corpus has shown that understanding both modalities is essential for detecting these techniques, which was further confirmed in our experiments with several state-of-the-art multimodal models.

In future work, we plan to extend the dataset in size, including with memes in other languages. We further plan to develop new multi-modal models, specifically tailored to fine-grained propaganda detection, aiming for deeper understanding of the semantics of the meme and of the relation between the text and the image. A number of promising ideas have been already tried by the participants in a shared task based on this data at SemEval-2021 \cite{SemEval2021:task6}, which can serve as an inspiration when developing new models.

\section*{Ethics and Broader Impact}
\paragraph{User Privacy}

Our dataset only includes memes and it does not contain any user information.

\paragraph{Biases}

Any biases found in the dataset are unintentional, and we do not intend to do harm to any group or individual. We note that annotating propaganda techniques can be subjective, and thus it is inevitable that there would be biases in our gold-labeled data or in the label distribution. We address these concerns by collecting examples from a variety of users and groups, and also by following a well-defined schema, which has clear definitions. Our high inter-annotator agreement makes us confident that the assignment of the schema to the data is correct most of the time.

\paragraph{Misuse Potential}

We ask researchers to be aware that our dataset can be maliciously used to unfairly moderate memes based on biases that may or may not be related to demographics and other information within the text. Intervention with human moderation would be required in order to ensure this does not occur.

\paragraph{Intended Use}

We present our dataset to encourage research in studying harmful memes on the web. We believe that it represents a useful resource when used in the appropriate manner. 

\section*{Acknowledgments}

This research is part of the Tanbih mega-project,\footnote{\url{http://tanbih.qcri.org/}} which is developed at the Qatar Computing Research Institute, HBKU, and aims to limit the impact of ``fake news,'' propaganda, and media bias by making users aware of what they are reading.

%% file: sections/supplemental_material.tex
\section{Annotation Instructions}
\label{sec:appendix_annotation_instructions}

\subsection{Guidelines for Annotators - Phases 1}
The annotators were presented with the following guidelines during phase 1 for filtering and editing the text of the memes. 

\subsubsection{Choice of memes/Filtering Criteria}
\label{apendix:rejection}

In order to ensure consistency for our data, we defined \emph{meme} as a photograph-style image with a short text on top. We asked the annotators to exclude memes with the below characteristics. During this phase, we filtered out 111 memes.

\begin{itemize}
    \itemsep0em
    \item Images with diagrams/graphs/tables.
    \item Memes for which no multimodal analysis is possible: e.g., only text, only image, etc.
    \item Cartoons.
\end{itemize}

\subsubsection{Rules for Text Editing}
\label{apendix:editing}
We used the Google Vision API\footnote{\url{http://cloud.google.com/vision}} to extract the text from the memes. As the output of the system sometimes contains errors, a manual checking was needed. Thus, we defined several text editing rules as listed below, and we applied them to the textual content extracted from each meme.

\begin{enumerate}
\itemsep0em
\item When the meme is a screenshot of a social network account, e.g., WhatsApp, the user name and login can be removed as well as all \emph{Like}, \emph{Comment}, and \emph{Share} elements.
\item Remove the text related to logos that are not part of the main text.
\item Remove all text related to figures and tables.
\item Remove all text that is partially hidden by an image, so that the sentence is almost impossible to read.
\item Remove text that is not from the meme, but on banners and billboards carried on by demonstrators, street advertisements, etc.
\item Remove the author of the meme if it is signed.
\item If the text is in columns, first put all text from the first column, then all text from the next column, etc.
\item Rearrange the text, so that there is one sentence per line, whenever possible.
\item If there are separate blocks of text in different locations of the image, separate them by a blank line. However, if it is evident that text blocks are part of a single sentence, keep them together.
\end{enumerate}

\begin{figure*}[t]
\centering
\includegraphics[width=0.8\textwidth]{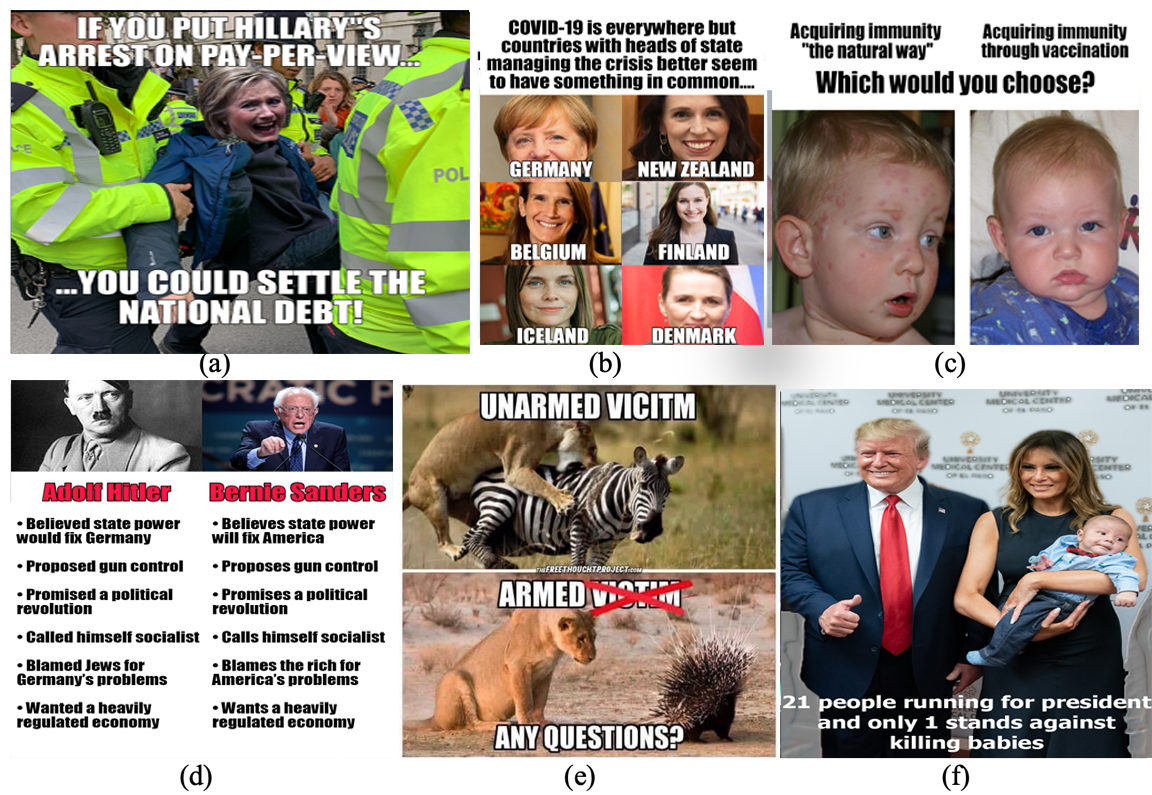}
\caption{Memes from our dataset. \textbf{Image sources:} \imgsrc{https://unsplash.com/photos/kBRrav94tGg}{(a) 1}, 
\imgsrc{https://commons.wikimedia.org/wiki/File:Hillary_Clinton_(25055755464).jpg}{(a) 2};
\imgsrc{https://upload.wikimedia.org/wikipedia/commons/b/bf/Angela_Merkel._Tallinn_Digital_Summit.jpg}{(b) 1},
\imgsrc{https://upload.wikimedia.org/wikipedia/commons/6/6c/Jacinda_Ardern\%2C_2018.jpg}{(b) 2},
\imgsrc{https://upload.wikimedia.org/wikipedia/commons/e/ed/SophieWilm\%C3\%A9s_beyear2020.jpg}{(b) 3},
\imgsrc{https://upload.wikimedia.org/wikipedia/commons/4/4c/Prime_Minister_of_Finland_Sanna_Marin_2019.jpg}{(b) 4},
\imgsrc{https://upload.wikimedia.org/wikipedia/commons/f/f2/Katr\%C3\%ADn_Jakobsd\%C3\%B3ttir_\%2824539871465\%29_\%28cropped\%29.jpg}{(b) 5},
\imgsrc{https://upload.wikimedia.org/wikipedia/commons/5/5d/Mette_Frederiksen_2019.jpg}{(b) 6};
\imgsrc{https://commons.wikimedia.org/wiki/File:Baby_Alfalfa.JPG}{(c) 1},
\imgsrc{https://commons.wikimedia.org/wiki/File:Oh_no_-_chicken_pox!_(486197031).jpg}{(c) 2};
\imgsrc{https://commons.wikimedia.org/wiki/File:Bernie_Sanders_(48023129387).jpg}{(d) 1},
\imgsrc{https://commons.wikimedia.org/wiki/File:Adolf-hitler-1.jpg}{(d) 2};
\imgsrc{https://i.imgur.com/Cyq7kdu.jpeg}{(e)};
\imgsrc{https://commons.wikimedia.org/wiki/File:President_Trump_and_the_First_Lady_in_El_Paso,_Texas_(48490970081).jpg}{(f)}; \textbf{Licenses:} \public{(b) 6, (c) 1, (f)}; \ccsnd{(a) 2,(b) 1, (b) 5, (c) 2, (d) 1}; \unplash{(a) 1}; \ccfrth{(b) 2, (b) 3, (b) 4, (c) 2}; \imgur{(e)}
}
\label{fig:meme_example_appendix}
\end{figure*}

\subsection{Guidelines for Annotators - Phases 2-5}
\label{ssec:appendix_guideline_phase2-5}
The annotators were presented with the following guidelines. In these phases, the annotations were performed by three annotators.

\subsubsection{Annotation Phase 2} 
Given the list of propaganda techniques for the text-only annotation task, as described in Section~\ref{ssec:appendix_definitions_prop_tech} (techniques 1-20), and the textual content of a meme, the task is to identify which techniques appear in the text and the exact  span for each of them. 

\subsubsection{Annotation Phase 4}

In this phase, the task was to identify which of the 22 techniques, described in Section~\ref{ssec:appendix_definitions_prop_tech}, appear in the meme, i.e.,~both in the text and in the visual content. Note that some of the techniques occurring in the text might be identified only in this phase because the image provides a necessary context. 

\subsubsection{Consolidation (Phase 3 and 5)} 
In this phase, the three annotators met together with other consolidators and discussed each annotation, so that a consensus on each of them is reached. These phases are devoted to checking existing annotations. However, when a novel instance of a technique is observed during the consolidation, it is added.

\subsection{Definitions of Propaganda Techniques}
\label{ssec:appendix_definitions_prop_tech}

\textbf{1. Presenting irrelevant data (Red Herring)}
Introducing irrelevant material to the issue being discussed, so that everyone's attention is diverted away from the points made.
\vspace{3pt}
\\Example 1: In politics, defending one's own policies regarding public safety -- ``\emph{I have worked hard to help eliminate criminal activity. What we need is economic growth that can only come from the hands of leadership.}''
\\Example 2: ``\emph{You may claim that the death penalty is an ineffective deterrent against crime -- but what about the victims of crime? How do you think surviving family members feel when they see the man who murdered their son kept in prison at their expense? Is it right that they should pay for their son's murderer to be fed and housed?}''

\paragraph{2. Misrepresentation of someone's position (Straw Man)}
When an opponent's proposition is substituted with a similar one, which is then refuted in place of the original proposition.
\vspace{3pt}
\\Example: \\\emph{Zebedee: What is your view on the Christian God?}\\
\emph{Mike: I don't believe in any gods, including the Christian one.}\\
\emph{Zebedee: So you think that we are here by accident, and all this design in nature is pure chance, and the universe just created itself?}\\
\emph{Mike: You got all that from me stating that I just don't believe in any gods?}

\paragraph{3. Whataboutism}
A technique that attempts to discredit an opponent's position by charging them with hypocrisy without directly disproving their argument.
\vspace{3pt}
\\Example 1: \textit{a nation deflects criticism of its recent human rights violations by pointing to the history of slavery in the United States.}
\\Example 2:\textit{``Qatar spending profusely on Neymar, not fighting terrorism''}

\paragraph{4. Causal oversimplification}
Assuming a single cause or reason when there are actually multiple causes for an issue.
It includes transferring blame to one person or group of people without investigating the complexities of the issue. An example is shown in Figure \ref{fig:meme_example_appendix}(b).
\vspace{12pt}
\\Example 1: ``\emph{President Trump has been in office for a month and gas prices have been skyrocketing. The rise in gas prices is because of him.}''
\\Example 2: \emph{The reason New Orleans was hit so hard with the hurricane was because of all the immoral people who live there.}

\paragraph{5. Obfuscation, Intentional vagueness, Confusion}
Using words which are deliberately not clear so that the audience may have their own interpretations. For example, when an unclear phrase with multiple definitions is used within the argument and, therefore, it does not support the conclusion.
\vspace{3pt}
\textit{~\\Example}: \emph{It is a good idea to listen to victims of theft. Therefore if the victims say to have the thief shot, then you should do that.}

\paragraph{6. Appeal to authority}
Stating that a claim is true simply because a valid authority or expert on the issue said it was true, without any other supporting evidence offered. We consider the special case in which the reference is not an authority or an expert in this technique, although it is referred to as \textit{Testimonial} in literature.
\vspace{3pt}
\\Example 1: \emph{Richard Dawkins, an evolutionary biologist and perhaps the foremost expert in the field, says that evolution is true. Therefore, it's true.}
\\Example 2: ``\emph{According to Serena Williams, our foreign policy is the best on Earth. So we are in the right direction.}''

\paragraph{7. Black-and-white Fallacy}
Presenting two alternative options as the only possibilities, when in fact more possibilities exist. We include \emph{dictatorship}, which happens when we leave only one possible option, i.e.,~when we tell the audience exactly what actions to take, eliminating any other possible choices. An example of this technique is shown in Figure~\ref{fig:meme_example_appendix}(c).
\vspace{3pt}
\\Example 1: \emph{You must be a Republican or Democrat. You are not a Democrat. Therefore, you must be a Republican.}
\\Example 2: \emph{I thought you were a good person, but you weren't at church today.}

\paragraph{8. Name Calling or Labeling}
Labeling the object of the propaganda campaign as either something the target audience fears, hates, finds undesirable or loves, praises.
\vspace{3pt}
\\Examples: \emph{Republican congressweasels}, \emph{Bush the Lesser}. Note that here \textit{lesser} does not refer to \emph{the second}, but it is \textit{pejorative}. 

\paragraph{9. Loaded Language} 
Using specific words and phrases with strong emotional implications (either positive or negative) to influence an audience.
\vspace{3pt}
\\Example 1: ``\emph{[...] a lone lawmaker’s childish shouting.}''
\\Example 2: ``\emph{how stupid and petty things have become in Washington.}''

\paragraph{10. Exaggeration or Minimisation}
Either representing something in an excessive manner: making things larger, better, worse (e.g.,~\emph{the best of the best}, \emph{quality guaranteed}) or making something seem less important or smaller than it really is (e.g.,~saying that an insult was just a joke). An example meme is shown in Figure~\ref{fig:meme_example_appendix}(a).
\vspace{3pt}
\\Example 1: ``\emph{Democrats bolted as soon as Trump’s speech ended in an apparent effort to signal they can't even stomach being in the same room as the President.}''
\\Example 2: ``\emph{We're going to have unbelievable intelligence.}''

\paragraph{11. Flag-waving}
Playing on strong national feeling (or to any group, e.g.,~race, gender, political preference) to justify or promote an action or idea.
\vspace{3pt}
\\Example 1: ``\emph{patriotism mean no questions}'' (this is also a slogan)
\\Example 2: ``\emph{Entering this war will make us have a better future in our country.}''

\paragraph{12. Doubt}
Questioning the credibility of someone or something.
\vspace{3pt}
\\Example: \emph{A candidate talks about his opponent and says: ``Is he ready to be the Mayor?''}

\paragraph{13. Appeal to fear/prejudice} 
Seeking to build support for an idea by instilling anxiety and/or panic in the population towards an alternative. In some cases the support is built based on preconceived judgements. An example is shown in Figure~\ref{fig:meme_example_appendix}(c).
\vspace{3pt}
\\Example 1: ``\emph{Wither we go to war or we will perish.}'' Note that, this is also a Black and White fallacy.
\\Example 2: ``\emph{We must stop those refugees as they are terrorists.}''

\paragraph{14. Slogans}
A brief and striking phrase that may include labeling and stereotyping. Slogans tend to act as emotional appeals.
\vspace{3pt}
\\Example 1: ``\emph{The more women at war\ldots the sooner we win.}''
\\Example 2: ``\emph{Make America great again!}''

\paragraph{15. Thought-terminating clich\'e}
Words or phrases that discourage critical thought and meaningful discussion about a given topic. They are typically short, generic sentences that offer seemingly simple answers to complex questions or that distract attention away from other lines of thought.
\vspace{3pt}
Examples: \emph{It is what it is; It's just common sense; You gotta do what you gotta do; Nothing is permanent except change; Better late than never; Mind your own business; Nobody's perfect; It doesn't matter; You can't change human nature.}

\paragraph{16. Bandwagon}
Attempting to persuade the target audience to join in and take the course of action because ``everyone else is taking the same action''.
\vspace{3pt}
\\Example 1: \emph{Would you vote for Clinton as president? 57\% say ``yes.''}
\\Example 2: \emph{90\% of citizens support our initiative. You should.}

\paragraph{17. Reductio ad hitlerum}
Persuading an audience to disapprove an action or idea by suggesting that the idea is popular with groups hated in contempt by the target audience. It can refer to any person or concept with a negative connotation. An examples is shown in Figure \ref{fig:meme_example_appendix}(d). 
\vspace{3pt}
\\Example 1: ``\emph{Do you know who else was doing that? Hitler!}''
\\Example 2: ``\emph{Only one kind of person can think in that way: a communist.}''

\paragraph{18. Repetition}
Repeating the same message over and over again so that the audience will eventually accept it.

\paragraph{19. Smears} 
A smear is an effort to damage or call into question someone's reputation, by propounding negative propaganda. It can be applied to individuals or groups. An example meme is shown in Figure \ref{fig:meme_example_appendix}(a).

\paragraph{20. Glittering generalities}
These are words or symbols in the value system of the target audience that produce a positive image when attached to a person or issue. Peace, hope, happiness, security, wise leadership, freedom, ``The Truth'', etc. are virtue words. Virtue can be also expressed in images, where a person or an object is depicted positively. In Figure \ref{fig:meme_example_appendix}(f), we provide an example to depict such a scenario. 

\paragraph{21. Transfer}
Also known as \emph{association}, this is a technique of projecting positive or negative qualities (praise or blame) of a person, entity, object, or value onto another to make the second more acceptable or to discredit it. It evokes an emotional response, which stimulates the target to identify with recognized authorities. Often highly visual, this technique often uses symbols (e.g.,~the swastikas used in Nazi Germany, originally a symbol for health and prosperity) superimposed over other visual images.

\paragraph{22. Appeal to (strong) emotions} 
Using images with strong positive/negative emotional implications to influence an audience. Figure \ref{fig:meme_example_appendix}(f) shows an example.

\section{Hyper-parameter Values}

In this section, we list the values of the hyper-parameters we used when training our models.
\begin{itemize}
    \item Batch size: 32
    \item Optimizer: AdamW
    \begin{itemize}
        \item Learning rate: 5e-5
        \item epsilon: 1e-8
        \item weight decay: 0.01
    \end{itemize}
    \item Max sequence length: 128
    \item Number of epochs: 37
    \item Early stopping: F1-micro on dev set
\end{itemize}

We further give statistics about the number of parameters for each model, so that one can get an idea about their complexity:
\begin{itemize}
    \item \textbf{ResNet-152}: 60,300,000
    \item \textbf{fastText}: 6,020 
    \item \textbf{BERT (bert-base-uncased)}: 110,683,414
    \item \textbf{fastText + ResNet-152 (early fusion)}: 11,194,398
    \item \textbf{BERT + ResNet-152 (late fusion)}: 170,983,752
    \item \textbf{MMBT}: 110,683,414
    \item \textbf{ViLBERT CC}: 112,044,290
    \item \textbf{VisualBERT COCO}: 247,782,404
\end{itemize}